\documentclass[runningheads]{template/llncs}
\usepackage{graphicx, amsmath, amssymb, color}
\usepackage{subfig, microtype, mwe, booktabs, pifont, arydshln, url, appendix}
\usepackage[dvipsnames]{xcolor}
\usepackage[acronym]{glossaries}
\usepackage{myAcronym}
\usepackage{definitions}
\usepackage[dvipsnames]{xcolor}

\arxivcopy % Note: Arxiv copy includes everything finalcopy does as well.

\begin{reviewonly}
    
\end{reviewonly}
\begin{finalonly}
    
\end{finalonly}

\begin{reviewonly}
    \usepackage{template/ruler}
    \usepackage[width=122mm,left=12mm,paperwidth=146mm,height=193mm,top=12mm,paperheight=217mm]{geometry}
\end{reviewonly}
\begin{arxivonly}
    \usepackage[width=122mm,left=12mm,paperwidth=146mm,height=193mm,top=12mm,paperheight=217mm]{geometry}
    \usepackage[pagebackref=true,breaklinks=true,colorlinks,bookmarks=false,linkcolor=NavyBlue]{hyperref}
\end{arxivonly}

\def\ECCVSubNumber{849}

\begin{arxivonly}
    \usepackage{times}
\end{arxivonly}

\begin{document}

\pagestyle{headings}
\mainmatter

%%%%%%%%%%%% TITLE %%%%%%%%%%%%

\title{Vision-Aided Radio: User Identity Match in Radio and Video Domains Using Machine Learning}

%%%---------- Review Authors ----------%%
%\begin{reviewonly}
%	\titlerunning{ECCV-20 submission ID \ECCVSubNumber} 
%	\authorrunning{ECCV-20 submission ID \ECCVSubNumber} 
%	\author{Anonymous ECCV submission}
%	\institute{Paper ID \ECCVSubNumber}
%\end{reviewonly}

%%---------- Final Authors ----------%%
\begin{finalonly}
	\titlerunning{Vision-Aided Radio}
	\author{Vinicius M. de Pinho$^1$ \and %\orcidID{0000-0003-0223-3599}
			Marcello L. R. De Campos$^1$  \and %\orcidID{0000-0001-8064-3054}
			Luis U. Garcia$^2$ \and %\orcidID{0000-0001-7016-4252}
			Dalia Popescu$^2$} %\orcidID{0000-0003-1971-1606}
	\authorrunning{V. Mesquita de Pinho et al.}
	\institute{$^1$Federal University of Rio de Janeiro \\ $^2$Nokia Bell Labs}
\end{finalonly}

\maketitle

%%%%%%%%%%%% ABSTRACT %%%%%%%%%%%%
\begin{abstract}
	5G is designed to be an essential enabler and a leading infrastructure provider in the communication technology industry by supporting the demand for the growing data traffic and a variety of services with distinct requirements. The use of deep learning and computer vision tools has the means to increase the environmental awareness of the network with information from visual data. Information extracted via computer vision tools such as user position, movement direction, and speed can be promptly available for the network. However, the network must have a mechanism to match the identity of a user in both visual and radio systems. This mechanism is absent in the present literature. Therefore, we propose a framework to match the information from both visual and radio domains. This is an essential step to practical applications of computer vision tools in communications. We detail the proposed framework training and deployment phases for a presented setup. We carried out practical experiments using data collected in different types of environments. This work compares the use of Deep Neural Network and Random Forest classifiers and shows that the former performed better across all experiments, achieving classification accuracy greater than $99\%$.
	
	%for improving models through in depth analysis across how many models and across how many datasets.
\end{abstract}

%%%%%%%%%%%% INTRO %%%%%%%%%%%%
%%%%%%%%%%%%%%%%%%%%%%%%%%%%%%%%%%%%%%%
% SECTION: INTRODUCTION
%%%%%%%%%%%%%%%%%%%%%%%%%%%%%%%%%%%%%%%
\section{Introduction}
\label{sec:intro}

5G systems and artificial intelligence (AI) have been highlighted as fields of innovation emblematic for the transition to a smarter society. Envisioned to offer a plethora of services and capabilities, 5G addresses a wide range of use cases, including enhanced mobile broadband, ultra-reliable low-latency communications, and massive machine-type traffic. 

Due to the advancements in AI techniques, especially deep learning, and the availability of extensive data, there has been an overwhelming interest in using AI for the improvement of wireless networks. Combining deep learning and \gls{cv} techniques have seen great success in diverse fields, such as security and healthcare, where they deliver state-of-the-art results in multiple tasks. Applying computer vision with deep learning in wireless communications has seen recent growing interest. Computer vision brings powerful tools to improve current communications systems. The use of visual information enriches the environmental awareness of networks and can enable context-aware communications to a level that is yet to be explored~\cite{tian2020applying}. 

Computer vision and deep learning have direct applications in the physical layer. We can exemplify an application with the following case. When using \gls{mimo} beamforming communication systems, beams' direction and power can be scheduled using the knowledge of users' locations and blocking cases readily available from the visual information. The immediate availability of data reduces overhead in communication, minimizing power consumption, and interference. Moreover, CV tools can give motion information about a user at the edge of the coverage area. This data can be used to project and estimate whether or when a terminal goes out or comes into its serving area. Then the network can allocate channel resources for the handover process to improve the utilization efficiency of the system resources.

In a practical scenario, visual data is acquired separately from radio information. It is only possible to take advantage of the ready-to-use visual information if the network can match the user identity from both visual and radio domains. Otherwise, the network does not have the means to use the information extracted from the visual data. The information from visual data that can be useful for the network, as in the following examples. For improving handover on edge cases by providing means of estimating a user's trajectories and speed; or reducing the radio control channel usage by contributing to user location instead of relying solely on radio information. To the best of our knowledge, a mechanism to match visual and radio data from the same user has not yet been described in the literature. The usual approach to deal with this problem is to consider only one user at a time in the scenario or to consider the information match is already provided for the network. Both do not happen in a realistic situation.

We close this gap by proposing a novel framework that enables the match of the user information from a visual-source with its radio counterpart. We model the problem as a classification task using the user position in the video feed and its \gls{cir}. We use a machine learning technique to solve the task of classifying the transmitting user. Our solution is a necessary step to allow the development of more complex scenarios involving the use of visual information in communications systems.  

The proposed framework is flexible; it is possible to incorporate as many users as necessary without critically increasing the computational complexity since the features used in the classification task are one-dimensional. Furthermore, we used an experimental setup to showcase the proposed framework. We carried out experiments using real data collected in four environments with different characteristics, from indoor spaces to an outdoor area. The high classification accuracy metrics in the experiments demonstrated the adaptability of the proposed framework. 

The industrial private networks can take great advantage of using the proposed framework. The industries' private networks require a customized design due to the strict requirements of ultra-reliable and low latency users and machine-type communications. There are numerous opportunities to explore in this environment, as flexibility increases. The operator owns both the \gls{ran} and the \gls{ue}; therefore, privacy becomes less of an issue. We have access to additional information to the \gls{ran}, data otherwise not available, for example, the video feed of the covered area. Hence, the network can extract useful information about the users, readily available on visual data, reducing the communication system's latency. 

%%%%%%%%%%%%%%%%%%%%%%%%%%%%%%%%%%%%%%%
% SUBSECTION: RELATED WORK
%%%%%%%%%%%%%%%%%%%%%%%%%%%%%%%%%%%%%%%
\subsection{Related Work}
\label{subsec:related}

Machine learning techniques have been used to solve various problems in communications systems. In \cite{Hoydis2017}, \cite{Simeone2018}, \cite{Zhang2019P}, \cite{ZhangBurg2020}, and \cite{valcarce2020joint}, some interesting use-cases of machine learning in the field of wireless communication and networking are surveyed: MAC layer protocols designed with reinforcement learning, deep neural networks for \gls{mimo} detection, \gls{ue} positioning with neural networks, and others. In \cite{valcarce2020joint}, the authors address the problem of designing signaling protocols for the MAC layer using reinforcement learning. The results show promising future for nonhuman-made protocols, they are faster and cheaper to construct when compared to the ones standardized by humans. Machine learning has been applied to \gls{mimo} detection, examples are the works with deep neural networks in \cite{goutay2020deep} and \cite{khani2019adaptive}. \gls{ue} positioning with neural networks as in \cite{Campos2014} and \cite{BUTT2020} can achieve mean positioning errors of less then $2$~m, essential for user localization in communication networks. Furthermore, machine learning-based solutions for communications can work with more than just radio signals to extend its capabilities. The use of computer vision-based on deep neural networks brings another source of useful tools.

Deep learning has succeed in the \gls{cv} field. The availability of large image and video datasets and high-speed affordable \glspl{gpu} has driven the researchers to develop deep-learning-based computer vision applications that excel in tasks such as image classification~\cite{he2015deep}, semantic segmentation~\cite{Goodfellow-et-al-2016}, and object detection~\cite{gonthier2020multiple}. Deep learning-based computer vision has been widely used in fields that generate a great number of visual data. Areas such as healthcare~\cite{liu2020deep}, remote sensing~\cite{li2020land}, and public security~\cite{matei2020deep}.

Recently, the scientific community started exploring the possibility of bringing intelligence from \gls{cv} systems to radio networks. In~\cite{alrabeiah2019viwi} the authors presented a framework for generating datasets with visual and radio information to facilitate research towards vision-aided wireless communication. The framework uses a game engine and a wireless propagation tool to recreate outdoor urban scenarios. This framework has been used for addressing beam-tracking and link-blockage problems.

The beam-tracking problem has been tackled in \cite{Alrabeiah2019} and also in \cite{tian2020applying}, using visual information from a dataset generated with the framework from~\cite{alrabeiah2019viwi}. The authors from \cite{alrabeiah2019viwi} combined images and beam indices from the scene generated by the framework to fine-tune a pre-trained deep learning model. However, the oversimplified scenario with only one user hinders the analysis if the method would scale to more complex scenarios.

The link-blockage problem was addressed in \cite{Alrabeiah2019} and \cite{charan2020visionaided}. The former tackles the problem in a reactive manner, i.e., the system classifies the present link status as blocked or not. The latter focuses on a proactive treatment of the problem, using recurrent neural networks to predict future blockage. Both works show promising results, but with only a single-moving user in the presence of stationary blockages.

The works in \cite{alrabeiah2019viwi}, \cite{Alrabeiah2019} and \cite{charan2020visionaided} can be further extended with more realistic scenarios. It is necessary to increase the number of possible users in the scene and allow non-stationary blockages. With a more dynamic scenario, the need to match the transmitting user in both video feed and radio transmission emerges. This issue is not addressed in \cite{alrabeiah2019viwi}, \cite{Alrabeiah2019} or \cite{charan2020visionaided}.

%%%%%%%%%%%%%%%%%%%%%%%%%%%%%%%%%%%%%%%%%%%%%%%%%%
% SUBSECTION: CONTRIBUTION AND PAPER ORGANIZATION
%%%%%%%%%%%%%%%%%%%%%%%%%%%%%%%%%%%%%%%%%%%%%%%%%%
\subsection{Contribution and Paper Organization}
\label{subsec:contribution}

We provide the possibility of user-identity matching in the radio domain and video domain by using machine learning.

Our contributions with this work are as follows:

\begin{itemize}
	\item We provide a general methodology that allows user-identity matching from radio and video domains using machine learning. The presented methodology is agnostic regarding the radio and video systems used or which machine learning technique is used for classification. In this sense, our methodology permits the incorporation of the best suitable technologies.
	\item Next, we showcase the proposed framework's feasibility, the steps for implementing and evaluating the proposed method and provide a detailed description of an experimental setup.
	\item We present and discuss results using Random Forest and Deep Neural Network classifiers on experimental data. We run practical experiments in four different environments and compare the classification results and training time. 
	
\end{itemize}
The paper is organized as follows.

\begin{itemize}
	\item Section \ref{sec:frameworkTestbed} describes the proposed framework and the testbed used throughout the paper. We start with the description of the testbed in Section~\ref{subsec:scenario} as it allows a more comprehensive and applied description of the framework. Section \ref{subsec:framework} describes the framework and methods for matching a \gls{ue} in a video feed to \gls{ue} identity in a radio transmission using machine learning and computer vision. The framework is described with a direct application on the testbed. 
	
	\item Experiments and results obtained in the testbed are detailed in Section~\ref{sec:experiments}. 
	
	\item Finally, conclusions are drawn in Section~\ref{sec:conclusions}.
\end{itemize}

%%%%%%%%%%%% FRAMEWORK %%%%%%%%%%%%
%%%%%%%%%%%%%%%%%%%%%%%%%%%%%%%%%%%%%%%%%%%%%%%%%%
% SECTION: FRAMEWORK AND TESTBED DESCRIPTION
%%%%%%%%%%%%%%%%%%%%%%%%%%%%%%%%%%%%%%%%%%%%%%%%%%
\section{Framework and Testbed Description}
\label{sec:frameworkTestbed}

%%%%%%%%%%%%%%%%%%%%%%%%%%%%%%%%%%%%%%%%%%%%%%%%%%
% SUBSECTION: EXPERIMENTAL SETUP DESCRIPTION
%%%%%%%%%%%%%%%%%%%%%%%%%%%%%%%%%%%%%%%%%%%%%%%%%%
\subsection{Experimental Setup Description}
\label{subsec:scenario}

In this section, we describe a simplified testbed that allows us to illustrate the principle of the proposed procedure, its feasibility, and how the experiments can be reproduced. We favored open software and communication entities, yet the concept can be extended to 5G devices for commercial use. 

\begin{figure}[h!]
	\centering
	\includegraphics[scale=1.0]{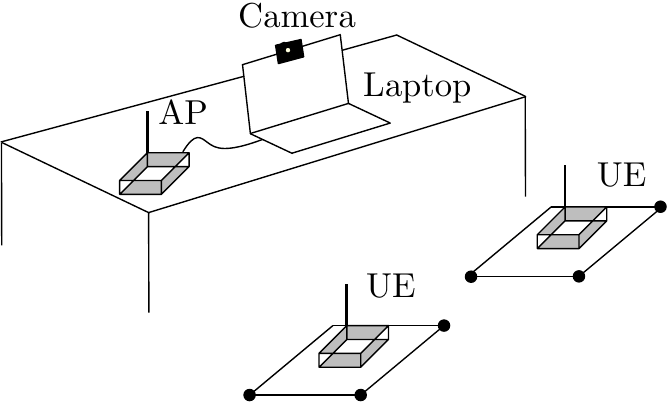}
	\caption{Setup for the testbed containing one camera, a laptop, and three USRPs.}
	\label{fig:setup_only}
\end{figure}

The setup for testbed is illustrated in Figure~\ref{fig:setup_only}. It consists of a \gls{gpu}-enabled laptop, a camera, an \gls{ap} and two identical, visually indistinguishable \glspl{ue}.

%%%%%%%%%%%%%
% FIGURE
%%%%%%%%%%%%%
%%%%%%%%%%%%%
% END FIGURE
%%%%%%%%%%%%%

The \gls{ap} and user devices are implemented using \glspl{usrp} model Ettus B210. We implement a simplified uplink transmission using GNU Radio~\cite{url:gnu-radio} based on the IEEE 802.11a \gls{ofdm} standard~\cite{terry2002ofdm}. The active user \gls{usrp} sends a pilot-based frame to the \gls{ap}. The frame uses a 52-subcarrier \gls{ofdm} operating at $1$~GHz. All the subcarriers are used to transmit pilots. The frame is modulated with a binary phase-shift keying modulation. The \gls{usrp} playing the \gls{ap} part is connected to the laptop, where the received signal is processed with GNU Radio.

The acquisition of the video stream is done with a Logitech C922 Pro Stream HD webcam, connected to the laptop. 

An equivalent 5G setup would have the following correspondence with our experimental setup. The \gls{ap} is the gNB and the two \glspl{ue} are the 5G User Devices (e.g., robots in industrial networks). The camera can be collocated with the gNB or the RAN can be connected through a communication interface to the camera. The processing done in the \gls{gpu} computer can be executed at the gNB site or other entity of the RAN (e.g., the RAN-LMF).

%%%%%%%%%%%%%%%%%%%%%%%%%%%%%%%%%%%%%%%%%%%%%%%%%%
% SUBSECTION: FRAMEWORK
%%%%%%%%%%%%%%%%%%%%%%%%%%%%%%%%%%%%%%%%%%%%%%%%%%
\subsection{Framework}
\label{subsec:framework}

\begin{figure}[ht]\centering
	\includegraphics[scale=0.8]{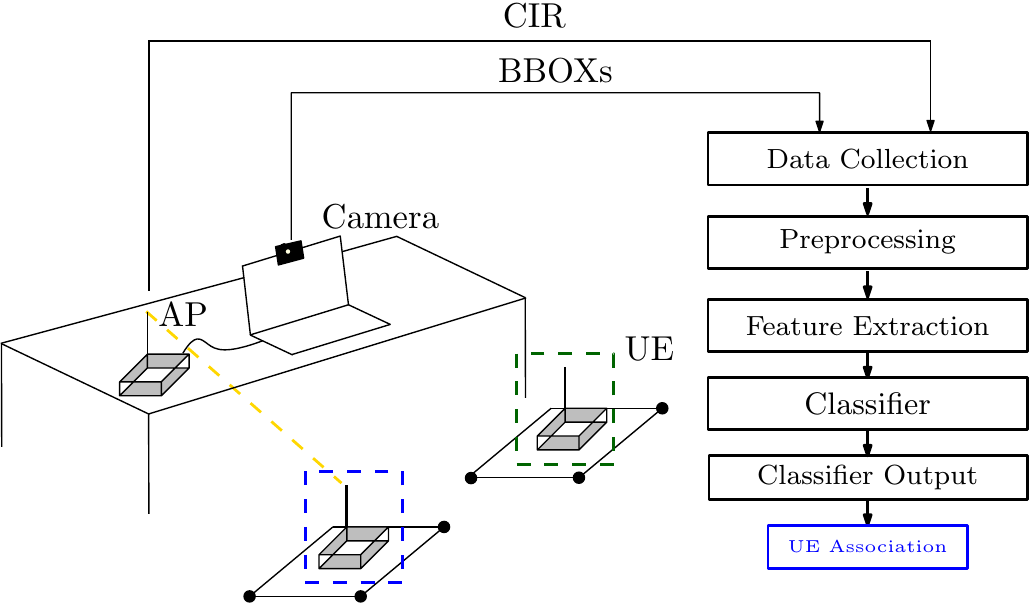}
	\caption{Illustration of framework steps linked with the experimental setup.}
	\label{fig:framework}
\end{figure}

We model the user-matching task as a classification problem and use a machine learning approach to solve it. The steps of the framework are visually illustrated in Figure~\ref{fig:framework} and summarized as follows.

\begin{itemize}
	\item Data collection: acquisition of data from the video system and the radio system;
	\item Preprocessing: merge of data from both sources and purge of spurious samples;
	\item Feature extraction: extraction of relevant features from preprocessed data;
	\item Training the ML model. In the following we will detail the option using Random Forest and Neural network classifiers: 
	\begin{itemize}
		\item Classifier: classification of input features;
		\item Classifier Output: label and level of confidence;
	\end{itemize}
	\item UE Association: association of classifier output with corresponding user information.
\end{itemize}

\pagebreak
%%%%%%%%%%%%%%%%%%%%%%%%%%%%%%%%%%%%%%%%%%%%%%%%%%
% SUBSECTION: DATA COLLECTION
%%%%%%%%%%%%%%%%%%%%%%%%%%%%%%%%%%%%%%%%%%%%%%%%%%
\subsection{Data Collection}
\label{subsec:data-coll}

\begin{figure}[h!]
	\centering
	\includegraphics[scale=0.6]{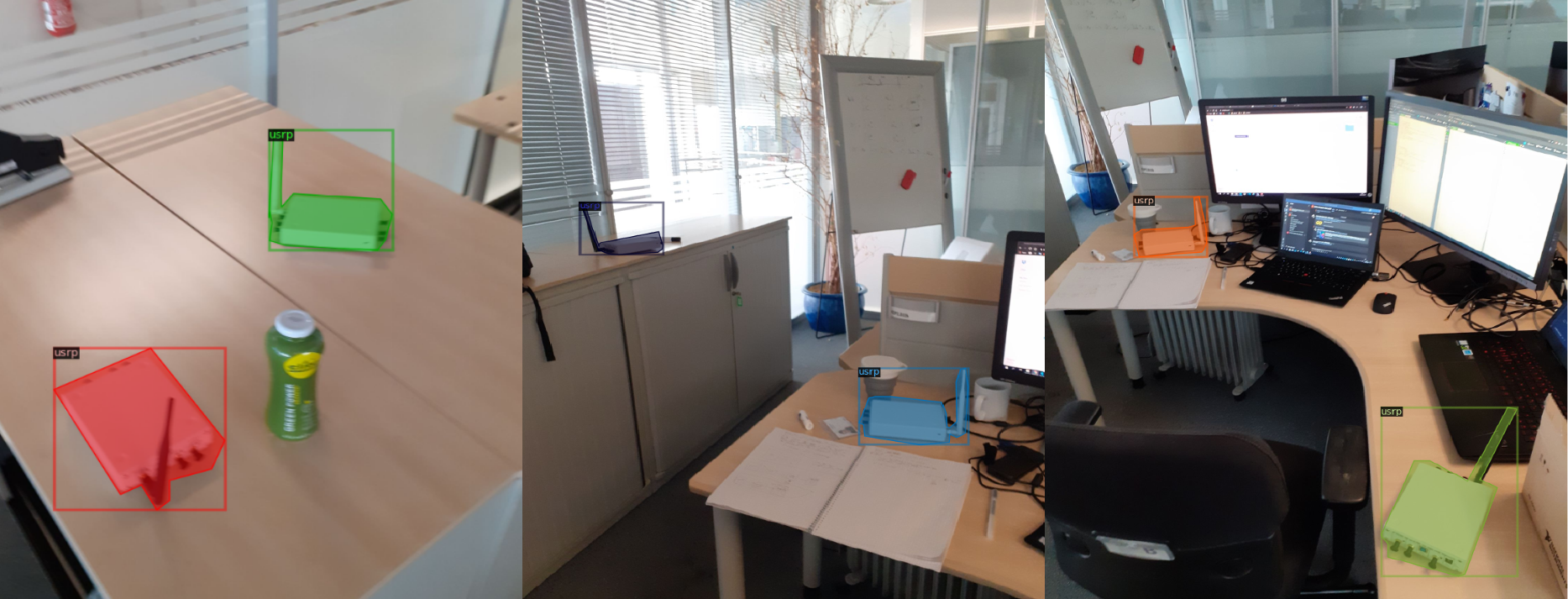}
	\caption{Examples of manually annotated images with bounding boxes around USRPs, used for fine-tuning the model pretrained on the COCO dataset.}
	\label{fig:bb_d2_all}
\end{figure}

The first essential step for the collection of video data is the recognition of the radio devices in the video feed.
Recognizing an object in a video feed is a well-known computer-vision task and we apply an existing ready-to-use framework to detect radio devices, in our case, \glspl{usrp}, in the video feed. We use and adapt an object-detection tool available in the Detectron2 framework~\cite{wu2019detectron2}. The tool is trained to recognize the devices by fine-tuning a mask region-based convolutional neural network that was pretrained on the COCO dataset
~\cite{lin2014microsoft}. Figure~\ref{fig:bb_d2_all} shows three examples of manually annotated images containing \glspl{usrp} with surrounding \glspl{bb} used to fine-tune the model. The reader is referred to~\cite{wu2019detectron2} for a complete description of the Detectron2 framework and means for fine-tuning to custom data. The output of the tool is an array with the \glspl{bb}, which indicates the radio devices' positions in the video feed. In addition, levels of confidence of the detection of the objects are provided. In summary, the data we collect from the video feed are the arrays with the \glspl{bb}, indicating the position of the devices in the scene, along with their levels of confidence of the detection.

%%%%%%%%%%%%%
% FIGURE
%%%%%%%%%%%%%
%%%%%%%%%%%%%
% END FIGURE
%%%%%%%%%%%%%

The space analyzed by the camera is limited to the area where the object detection is done with accuracy of $99$\% or higher, and the devices can move freely within the area. The high accuracy is imposed to avoid spurious measurements in the testbed.

The data collected from the radio system are the \glspl{cir}. The \gls{cir} is computed in GNU Radio with the pilot-based frames from the link between the transmitting device and the \gls{ap}. The set of \glspl{cir} computed during transmission is stored.

%%%%%%%%%%%%%%%%%%%%%%%%%%%%%%%%%%%%%%%%%%%%%%%%%%
% SUBSECTION: PREPROCESSING
%%%%%%%%%%%%%%%%%%%%%%%%%%%%%%%%%%%%%%%%%%%%%%%%%%
\subsection{Preprocessing}
\label{subsec:preproc}

During data collection, the information from the vision and radio systems are acquired concurrently. Each source of data saves the collected measurements with a unique timestamp. We create a unified representation using both vision and radio sources by matching their timestamps. 

With the measurements unified, the collected measurements are preprocessed. The \gls{cir} records with a maximum magnitude below a threshold $\delta$ are discarded. This is done because \glspl{cir} are wrongly estimated in the GNU Radio due to synchronization issues in a small number of transmitted frames. After this data-cleaning step, the remaining inputs are fed to the feature extractor.

For the training phase, the \glspl{bb} are coded into a label number, as illustrated in
Figure~\ref{fig:framework_details_train}. The vision system outputs a vector with \gls{bb} for each of the two devices presented in the scene. When there are two devices in the scene, one gets a \gls{bb} named ``\gls{bb} 1'' and the other the ``\gls{bb} 2''. Given that in our testbed there are only two devices, the following situations will be treated: when device ``\gls{bb} 1'' is transmitting and the one named  ``\gls{bb} 2'' is not the training label generated is $X = 1$. The training label $X = 2$ is generated when the device named ``\gls{bb} 2'' is transmitting and ``\gls{bb} 1'' is not. When no device is transmitting, the label generated is $X = 0$, also called  ``NO TX''.  Hence our system is going to be trained to classify three different situations, designed with the label $X \in \mathcal{X} = \{0, 1, 2\}$. 

In this work, we do not consider the case of two users transmitting simultaneously due to equipment limitations. However, the extension is straightforward when increasing the number of \glspl{ap} or using a user-multiplexing technique. Furthermore, for the practical experiments we carried out, the devices were moved throughout the setup area, and the system periodically reassessed the labels to the devices.  

%%%%%%%%%%%%%%%%%%%%%%%%%%%%%%%%%%%%%%%%%%%%%%%%%%
% SUBSECTION: FEATURE EXTRACTION
%%%%%%%%%%%%%%%%%%%%%%%%%%%%%%%%%%%%%%%%%%%%%%%%%%
\subsection{Feature extraction}
\label{subsec:feat-extr}

We identified the following features of the \gls{cir}, defined in~\eqref{eq:cir}, as being relevant for our problem: the \gls{cir} magnitude, phase, and the value and sample index of the \gls{cir} magnitude peak in the radio frame. 
\begin{equation}
\label{eq:cir}
h(t) = \sum_{k=0}^{N-1} a_{k} \mathrm{e}^{\mathrm{j}\theta_{k}}    \delta(t - \tau_{k}), 
\end{equation}
where $k$ is integer, $N$ is the number of multipath components, $a_{k}$, $\tau_{k}$, and $\theta_{k}$ are the random amplitude, propagation delay and phase of the $k$th multipath component, respectively. $\delta$ is the Dirac delta function.

\begin{figure}[h!]
	\centering
	\includegraphics[scale=0.7]{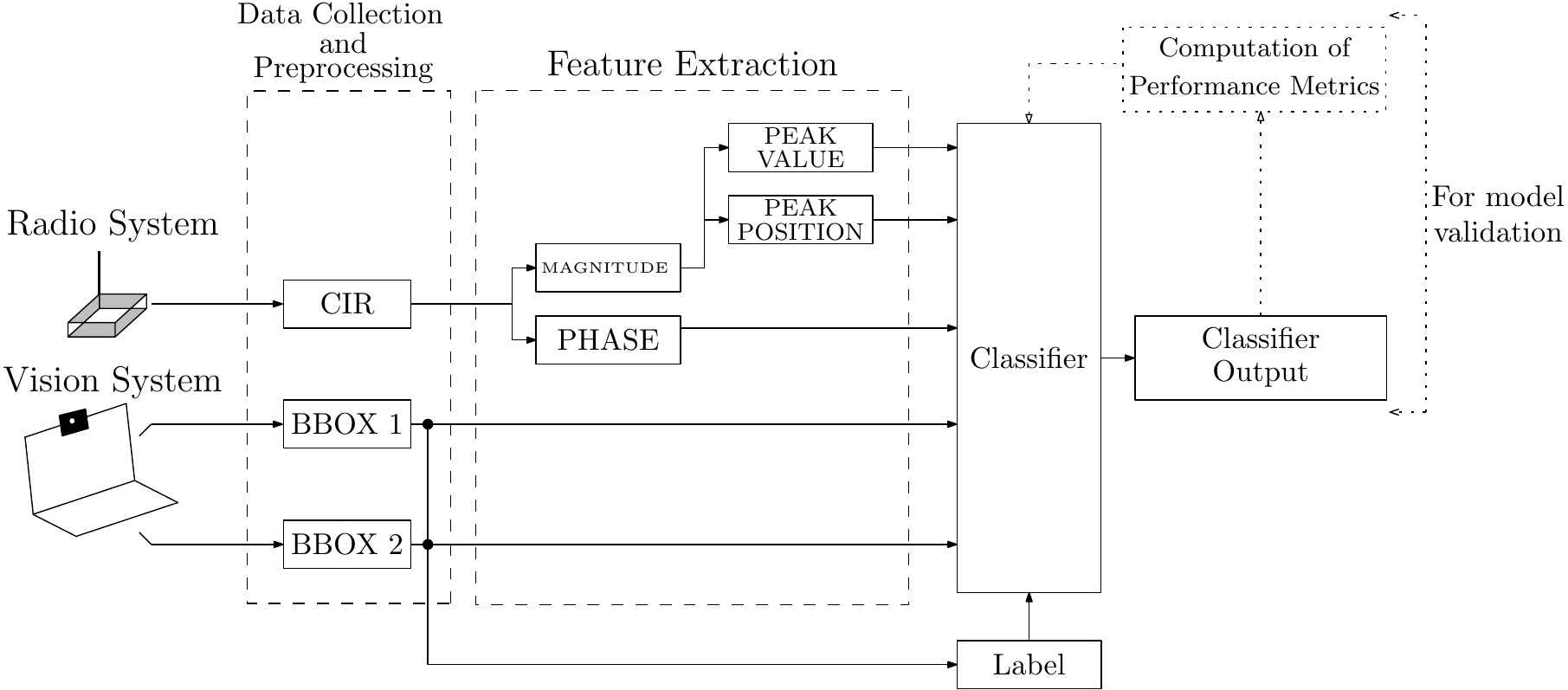}
	\caption{Details of the framework used for training and validation phases.}
	\label{fig:framework_details_train}
\end{figure}

From the vision system, we are using the array with the \glspl{bb}. Figure~\ref{fig:framework_details_train} shows the feature extraction steps in the framework used for training the model.

%%%%%%%%%%%%%
% FIGURE
%%%%%%%%%%%%%
%%%%%%%%%%%%%
% END FIGURE
%%%%%%%%%%%%%

%%%%%%%%%%%%%%%%%%%%%%%%%%%%%%%%%%%%%%%%%%%%%%%%%%
% SUBSECTION: RANDOM FOREST CLASSIFIER
%%%%%%%%%%%%%%%%%%%%%%%%%%%%%%%%%%%%%%%%%%%%%%%%%%
\subsection{Random Forest Classifier}
\label{subsec:rand-forest}

\begin{figure}[h!]
	\centering
	\includegraphics[scale=0.8]{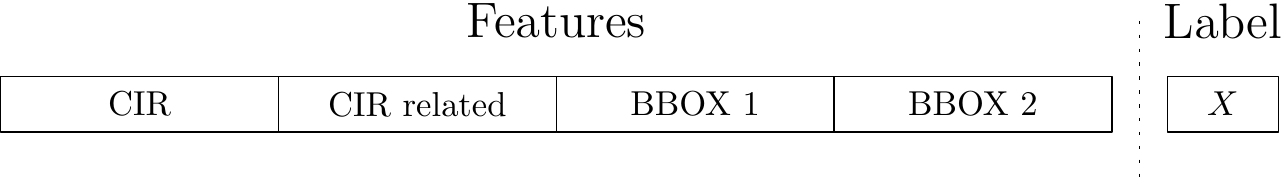}
	\caption{Input instance for the classifier, with the radio and video domain features and annotated with a label used for training and validation.}
	\label{fig:input_class}
\end{figure}

Figure~\ref{fig:input_class} shows the input for the classifier. The labels are used for supervised model training. Afterward, the trained model can be used in the deployment phase, as illustrated by the framework in Figure~\ref{fig:framework_details_deploy}, with only the features to classify new data. In this work, we train the models with \glspl{rfc} and \glspl{dnn}. The proposed framework is agnostic to the classifier used. We used \glspl{rfc} and \glspl{dnn} because both techniques are robust and give good classification results. 

%%%%%%%%%%%%%
% FIGURE
%%%%%%%%%%%%%
%%%%%%%%%%%%%
% END FIGURE
%%%%%%%%%%%%%

The \gls{rfc} is an ensemble learning algorithm for classification that uses decision trees~\cite{Denisko2018}. The \gls{rfc} constructs a large number of decision trees at training time and outputs the class that is the mode of the output from the individual trees. 

We train the model combining an exhaustive grid search over \gls{rfc} parameter values. The search space is confined to $20$--$50$ for the number of trees with a maximum depth between $30$ and $80$. The training uses $10$-fold cross-validation procedure, where the training dataset is split into $10$ smaller sets, the model is trained using $9$ of the folds and validated on the remaining part of the data. To evaluate the performance of the trained model, in each iteration we compute two different metrics: the logarithmic loss and the $F_{1}$ score. We choose the best model given the performance metrics. Furthermore, the best-trained model for a given dataset is used for testing, where we compute the confusion matrix, precision, recall, $F_{1}$ score, and classification accuracy. 

%%%%%%%%%%%%%%%%%%%%%%%%%%%%%%%%%%%%%%%%%%%%%%%%%%%
% SUBSECTION: CLASSIFIER OUTPUT AND UE ASSOCIATION
%%%%%%%%%%%%%%%%%%%%%%%%%%%%%%%%%%%%%%%%%%%%%%%%%%%
\subsection{Classifier Output and UE Association}
\label{subsec:clas-output-ue-assoc}

The classifier output is the predicted label number indicating which user is transmitting in the scene along with the level of confidence of the output. During the training procedure, the classifier output is used to compute the performance metrics, as illustrated in Figure~\ref{fig:framework_details_train} using dotted lines. 

\begin{figure}[h!]
	\centering
	\includegraphics[scale=0.65]{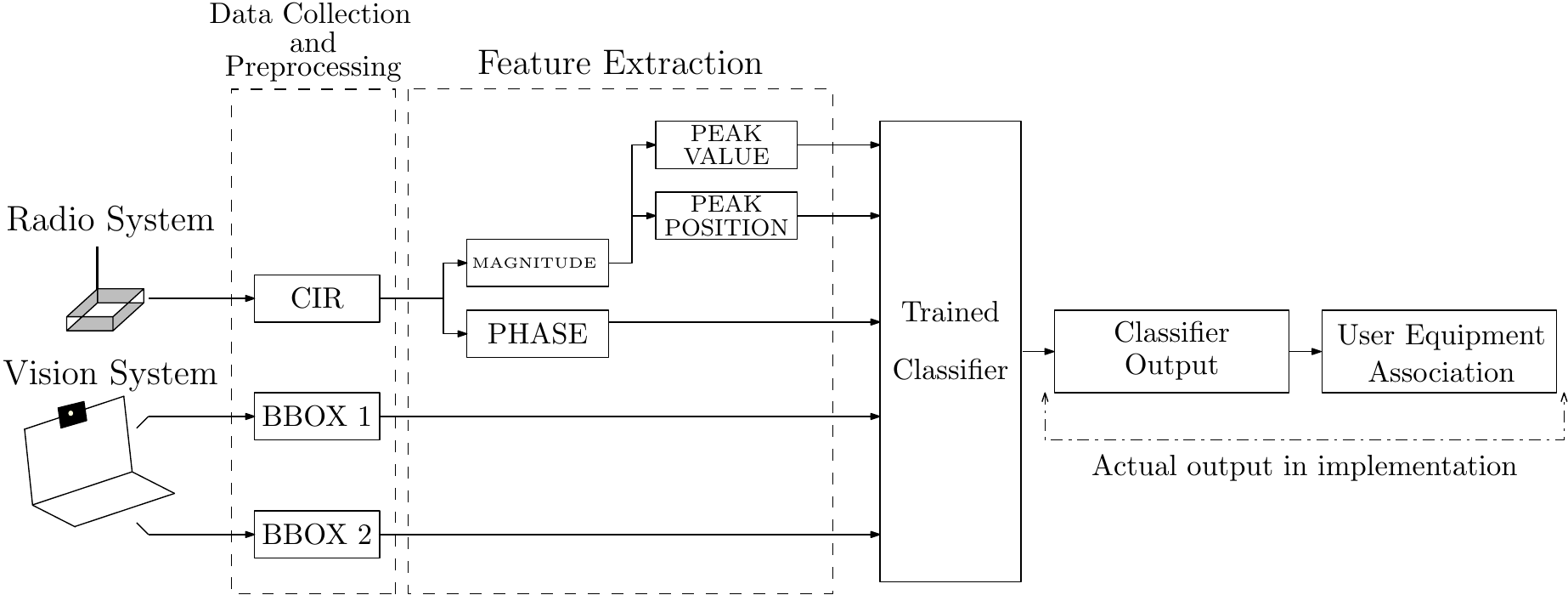}
	\caption{Details of the framework used for deployment.}
	\label{fig:framework_details_deploy}
\end{figure}

For deployment, the framework we use is shown in Figure~\ref{fig:framework_details_deploy}. The output is used to make the association with the device. When two possible users are in the scene, if the predicted label is $X = 1$, the device associated with the ``\gls{bb} $1$'' is the one transmitting in the scene, analogously for when the label is $X = 2$. When the predicted labels are $X = 0$, no user is transmitting to the \gls{ap} in the scene. In the scenario with only one user, the possible outcomes are: the predicted label is $X = 1$ when the user is transmitting, or $X = 0$ when no one is transmitting. With this step done, we have matched the information from both radio and video systems. In summary, the vision system detects two devices and is able to tell which one is transmitting, successfully matching visual and radio information.

%%%%%%%%%%%%%%%%%%%%%%%%%%%%%%%%%%%%%%%%%%%%%%%%%%%
% SUBSECTION: ALTERNATIVE ML SOLUTION
%%%%%%%%%%%%%%%%%%%%%%%%%%%%%%%%%%%%%%%%%%%%%%%%%%%
\subsection{Alternative ML Solution: Deep Neural Network Classifier}
\label{subsec:alt-ml-sol-nn}

The deep neural network classifier that we use is a feedforward neural network or multilayer perceptron. The architecture we use in this work is detailed in Table~\ref{tab:nn-arch}. The \gls{dnn} consists of an input layer, where the same input as the \gls{rfc} is used, followed by three hidden layers and an output layer. We use three hidden layers, each one with ReLu \cite{agarap2018deep} as activation function, followed by a dropout layer with rate of $0.5$, used to hinder overfitting. The output layer uses softmax as an activation function. During training, the labels are encoded using one-hot encoding to transform categorical data into a non-ordinal numerical representation. For details on the typical implementation of neural networks, the reader is referred to \cite{Goodfellow-et-al-2016} and \cite{aurelin2019}.

% Table with NN arch. description
% \usepackage{graphicx}
\begin{table}[!h]
\centering
\caption{Neural network architecture, with Specified Parameters for Each Layer and Number of Trainable Parameters. }
\vspace*{0.5cm}
\label{tab:nn-arch}
\resizebox{0.6\textwidth}{!}{%
\begin{tabular}{|l|l|l|l|}
\hline
\multicolumn{1}{|l|}{Layer} & \multicolumn{1}{l|}{Layer Type} & \multicolumn{1}{l|}{Parameters} & \multicolumn{1}{l|}{\# Parameters} \\ \hline
Layer 1 & Dense + ReLu    & Units: 256 & 15616 \\ \hline
Layer 2 & Dropout         & Rate: 0.5  & 0     \\ \hline
Layer 3 & Dense + ReLu    & Units: 128 & 32896 \\ \hline
Layer 3 & Dropout         & Rate: 0.5  & 0     \\ \hline
Layer 4 & Dense + ReLu    & Units: 64  & 8256  \\ \hline
Layer 5 & Dense + Softmax & Units: 3   & 195   \\ \hline
\end{tabular}%
}
\end{table}

%%%%%%%%%%%% EXPERIMENTS %%%%%%%%%%%%
%%%%%%%%%%%%%%%%%%%%%%%%%%%%%%%%%%%%%%%%%%%%%%%%%%%
% SECTION: EXPERIMENTS AND RESULTS
%%%%%%%%%%%%%%%%%%%%%%%%%%%%%%%%%%%%%%%%%%%%%%%%%%%
\section{Experiments and Results}
\label{sec:experiments}

We carried out experiments to evaluate the performance of the proposed framework in matching the correct users to their identities. Four experimental configurations with different dynamics were used, hence each set of measurements has distinct characteristics. This allowed us to test the capacity of our method to operate in different environments. 

Setup 1 was located in an indoor environment. An $18$~$m^{2}$ furnished room and only one person inside, to avoid fluctuations in the \gls{cir} measurements. For the measurement campaign, the equipment was put in place, as described in Section~\ref{subsec:scenario}. We defined an area of $2$~$m^{2}$ in front of the camera, where the user devices could move freely. The object-detection tool could survey the whole space and detect the devices with high accuracy, to avoid spurious measurements. We collected data for training and validation separately. The video and radio information was stored in the laptop's hard drive. For the measurements in this setup, there were $233{,}960$ instances collected. Being $176{,}874$ for training and $57{,}086$ for validation. The number of instances acquired during the measurement campaign is detailed in Table~\ref{tab:num-instances}.

Setup 2 was arranged in the corridor of office space. The environment has a different geometry than the other setup places. There are more reflections of the transmitted signal, which affects the \gls{cir} measurements. The setup place also tests the vision system ability to recognize the \glspl{usrp} in a different environment. The measurement campaign followed the same procedures as in Setup 1. In this case, a total of $397{,}073$ instances were collected. 

Setup 3 was placed inside a laboratory with electronic equipment. We followed the same steps for the measurement campaigns as the previous setups. The level of noise in the measurements was higher than in the previous experimental configurations. For this reason, the measurement campaign collected more data in this setup. Table~\ref{tab:num-instances} shows we acquired two times more instances in Setup 3 when compared to Setup 1.

\begin{figure}[h!]
	\centering
	\includegraphics[scale=0.3]{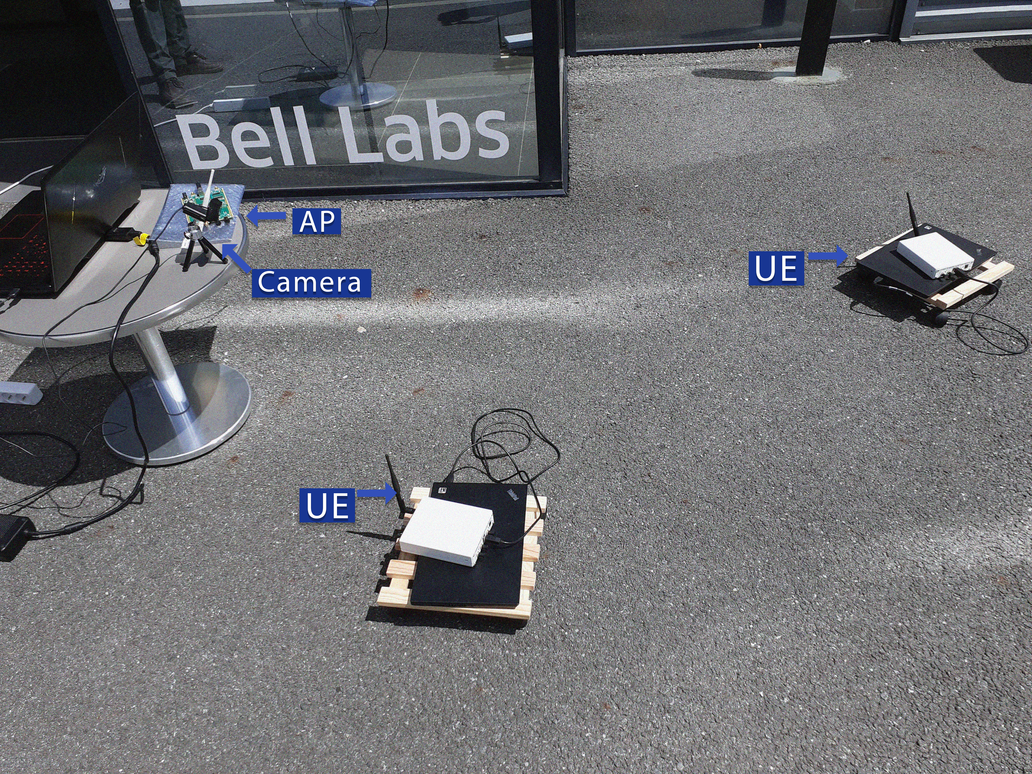}
	\caption{Setup for experiment 4, carried out in an outdoor area.}
	\label{fig:result_experiment4_image}
\end{figure}

Data collected for Setup 4 test our solution in an outdoor scenario. Setup 4, as shown in Figure~\ref{fig:result_experiment4_image}, was situated outside the building. The measurements done outdoors affect the \gls{cir} estimation. This brings different characteristics to the datasets acquired in this place. We followed the same steps for the measurement campaign as in the previous setups. For Setup 4 a total of $54{,}158$ instances were collected. They were $38{,}145$ for training and $16{,}013$ for validation.

%%%%%%%%%%%%%
% FIGURE
%%%%%%%%%%%%%
%%%%%%%%%%%%%
% END FIGURE
%%%%%%%%%%%%%

% Table with number of instance in each experiment
% \usepackage{graphicx}
\begin{table}[!h]
\centering
\caption{Number of Instances in the Training and Validation Datasets per Experiment.}
\label{tab:num-instances}
\vspace*{0.5cm}
\resizebox{0.5\textwidth}{!}{%
\begin{tabular}{l|l|l|l|}
\cline{2-4}
\multicolumn{1}{c|}{}              & \multicolumn{3}{c|}{Number of Instances} \\ \hline
\multicolumn{1}{|c|}{Setup} & \multicolumn{1}{c|}{Training} & \multicolumn{1}{c|}{Validation} & \multicolumn{1}{c|}{Total} \\ \hline
\multicolumn{1}{|l|}{Setup 1} & 176,874      & 57,086      & 233,960     \\ \hline
\multicolumn{1}{|l|}{Setup 2} & 242,975      & 154,098     & 397,073     \\ \hline
\multicolumn{1}{|l|}{Setup 3} & 380,527      & 105,187     & 485,714     \\ \hline
\multicolumn{1}{|l|}{Setup 4} & 38,145       & 16,013      & 54,158      \\ \hline
\end{tabular}%
}
\end{table}

%%%%%%%%%%%%%
% FIGURE
%%%%%%%%%%%%%
%%%%%%%%%%%%%
% END FIGURE
%%%%%%%%%%%%%

For each setup measurements, we preprocessed the training and validation data and extracted the engineered features, as detailed in Section~\ref{subsec:framework}. We carried out experiments using \glspl{rfc} and \gls{dnn} classifiers. The \gls{dnn} classifiers were trained during 10 epochs. The learning rate used was $0.001$. The architecture is presented in Table~\ref{tab:nn-arch}. The layers were initialized using the Glorot uniform initializer and no bias. Training for all the experiments was carried out in a Dell G3 3590 Laptop, with an Intel i7-9750H, 8~GB of RAM, and an NVIDIA GTX 1660 Ti Max-Q 6~GB. The training time is an average of running the same procedure $10$ times.

%%%%%%%%%%%%%%%%%%%%%%%%%%%%%%%%%%%%%%%%%%%%%%%%%%%
% SUBSECTION: PERFORMANCE METRICS
%%%%%%%%%%%%%%%%%%%%%%%%%%%%%%%%%%%%%%%%%%%%%%%%%%%
\subsection{Performance metrics}
\label{subsec:perf_metr}

We evaluated the trained models' performance in the classification task on the validation dataset. We plotted the confusion matrix. For easier comprehension, the labels defined in Section~\ref{subsec:preproc} are called by ``NO TX'', ``BBOX 1'' and ``BBOX 2'' for $X = 0$, $X = 1$ and $X = 2$, respectively. Furthermore, we compute the accuracy, average precision, recall, and $F_{1}$-score~\cite{Tharwat2018}. Accuracy is the percentage of the predicted outputs that exactly matched the corresponding set of true labels. Moreover, precision is computed as $tp/(tp + fp)$, where $tp$ is the number of true positives and $fp$ the number of false positives. The precision discloses the ability of the classifier not to label as positive a sample that is negative. Recall tells us the ability of the classifier to find all the positive samples. The recall score is computed as $tp/(tp + fn)$, where $fn$ is the number of false negatives. Furthermore, $F_{1}$-score is the harmonic mean of the precision and recall, it can be computed as $tp/(tp + 0.5[fp + fn])$. The highest possible value of the $F_{1}$-score is $1$, indicating perfect precision and recall, and the lowest possible value is $0$, if either the precision or the recall is zero. In this work, the $F_{1}$-score is obtained using the micro-averaging approach, i.e., we considered the class frequency for computing the $F_{1}$-score because we have a unbalanced training dataset with fewer instances with label $X = 0$. 

%%%%%%%%%%%%%%%%%%%%%%%%%%%%%%%%%%%%%%%%%%%%%%%%%%%
% SUBSECTION: RESULTS
%%%%%%%%%%%%%%%%%%%%%%%%%%%%%%%%%%%%%%%%%%%%%%%%%%%
\subsection{Results}
\label{subsec:results}

The first experiment was the one with Setup 1 using a random forest classifier. Training time took $12.21$ minutes. The validation results are the following. The accuracy was $94.09$\%, precision $0.96$, recall $0.96$, and $F_{1}$-score $0.96$. The confusion matrix is displayed in Figure~\ref{fig:cf_setup1_rfc}. From the confusion matrix, we can see that $11.7\%$ of the instances from ``BBOX 1'' were mistakenly classified as ``BOX 2''. The classifier assigns a wrong label to the validation dataset instance. This misclassification happens because the model is not able to differentiate the two users due to close positions of the devices in the video feed. Moreover, all the dataset instances with no device transmitting, labeled as ``NO TX'', were correctly classified. The dataset instances when no user is transmitting have null values in their fields, which makes it easy for the classifier to correctly label them.

The Setup 1 with neural network classifier took $03.50$ minutes to train. Figure~\ref{fig:cf_setup1_nn} displays the confusion matrix. The metrics computed show $99.91$\% of accuracy  precision of $0{.}99$, recall of $0{.}99$, and $F_{1}$-score of $0{.}99$. Therefore \gls{dnn} was not as prone to classification errors as \gls{rfc}.

The experiment with Setup 2 using the \gls{rfc} took $14.30$ minutes to train. The metrics results were: accuracy $99.77$\%, precision $0.99$, recall $0.99$, and $F_{1}$-score $0.99$. An equivalent analysis can be seen in the confusion matrix in Figure~\ref{fig:cf_setup2_rfc}. The confusion matrix shows that approximately $0.04\%$ (29 cases) of the instances from ``BBOX 1'' were misclassfied as ``BBOX 2''. For the instances labeled was ``BBOX 2'', only $0.48\%$ of the time the system incorrectly classified them as ``BBOX 1''.

For Setup 2 with neural network classifier, training time was $04.86$ minutes. The performance metrics were: accuracy  $99.98$\%, precision $0.99$, recall $0.99$ and $F_{1}$-score $0.99$. Figure~\ref{fig:cf_setup2_nn} shows the confusion matrix. In this case, only $19$ instances were incorrectly classified, which is negligible.

For the experiment on Setup 3 using the \gls{rfc} training time was $16.89$ minutes. The training duration was longer compared to the other experiments because the training dataset was the largest, as shown in Table~\ref{tab:num-instances}. For the validation dataset, the metrics are the following: accuracy $78.35$\%, precision $0.84$, recall $0.84$, and $F_{1}$-score $0.84$. The accuracy score is lower than the previous ones. However, the confusion matrix in Figure~\ref{fig:cf_setup3_rfc} shows that the system continues to perform well. It gets $100\%$ correct outputs when no device is transmitting in the scene. The instances with ``BOX 2'' were correctly classified with accuracy of $82\%$.

In the experiment in Setup 3 using a neural network, the training was $06.15$ minutes long. The confusion matrix for validation is displayed in Figure~\ref{fig:cf_setup3_nn}. The neural network classifier was able to handle the measurements in this setup better then the random forest due to the network's architecture capacity of generalization. The accuracy for this experiment was $99.76\%$. Precision, recall and $F_{1}$-score were all $0.98$. This shows the robustness of the neural network with the architecture presented in Table~\ref{tab:nn-arch}.   
Moreover, an experiment using Setup 4 was carried out using \gls{rfc}. The training time of $06.10$ minutes. The measurement campaign for Setup 4 was shorter, leading to smaller training and validation datasets. However, the system achieved great results as the metrics show. The accuracy was $99.66$\%. Precision was $0.99$, the same results for recall and $F_{1}$-score. The confusion matrix is shown in Figure~\ref{fig:cf_setup4_rfc}. 

The experiment with Setup 4 measurements using a neural network classifier had a training time of $02.01$ minutes. The confusion matrix is in Figure~\ref{fig:cf_setup4_nn}. Accuracy $99.99$\%, precision, recall, and $F_{1}$-score were $0.99$.

%%%%%%%%%%%%%%%%%%%%%%%%%%%%%%%%%%%%%%%
% DEFINE SIZE OF CONFUSION MATRIX PLOT
\newcommand{\sizeConfusionMatrix}{0.16}
%%%%%%%%%%%%%%%%%%%%%%%%%%%%%%%%%%%%%%%

%%%%%%%%%%%%%
% FIGURE
%%%%%%%%%%%%%
\begin{figure}[h!]
	\centering
	\includegraphics[scale=\sizeConfusionMatrix]{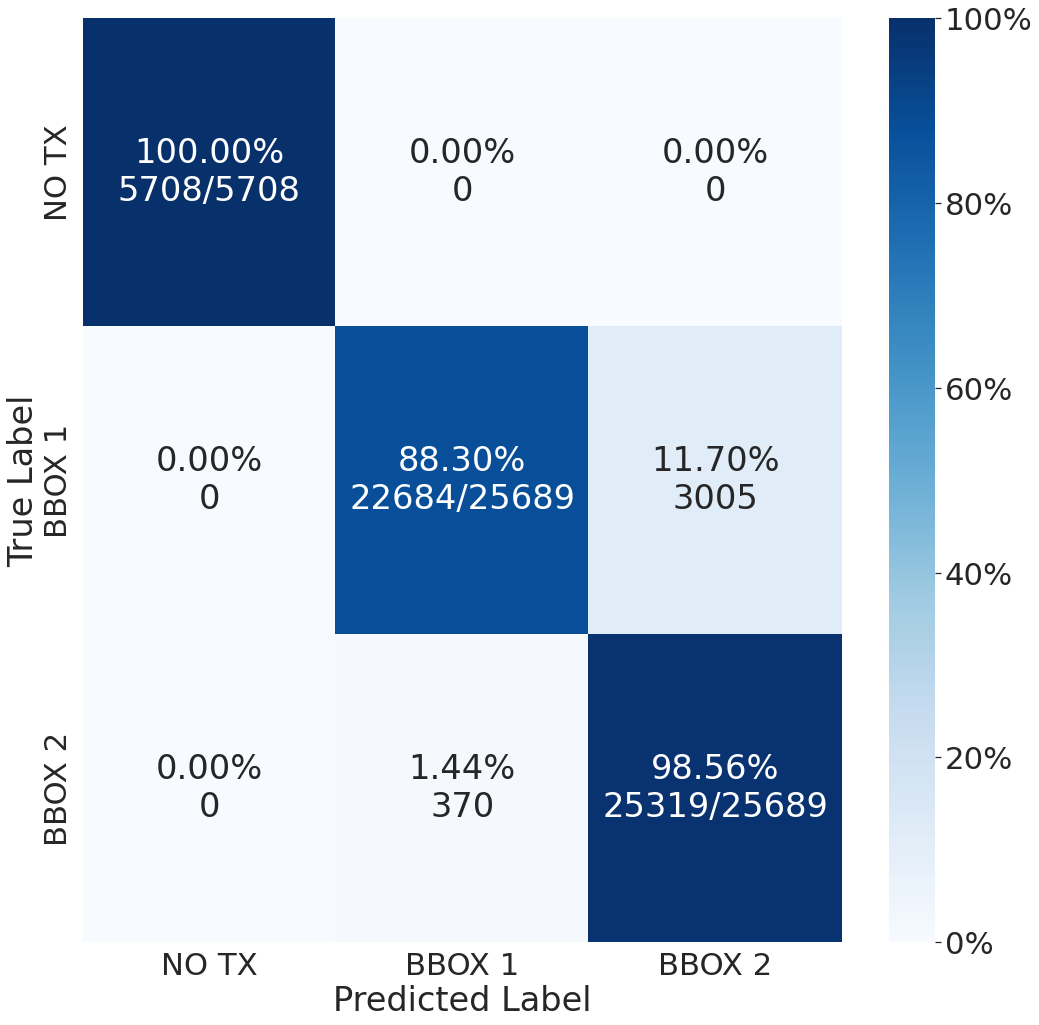}
	\caption{Confusion Matrix for Setup 1 data trained with Random Forest Classifier.}
	\label{fig:cf_setup1_rfc}
\end{figure}
%%%%%%%%%%%%%
% END FIGURE
%%%%%%%%%%%%%

%%%%%%%%%%%%%
% FIGURE
%%%%%%%%%%%%%
\begin{figure}[h!]
	\centering
	\includegraphics[scale=\sizeConfusionMatrix]{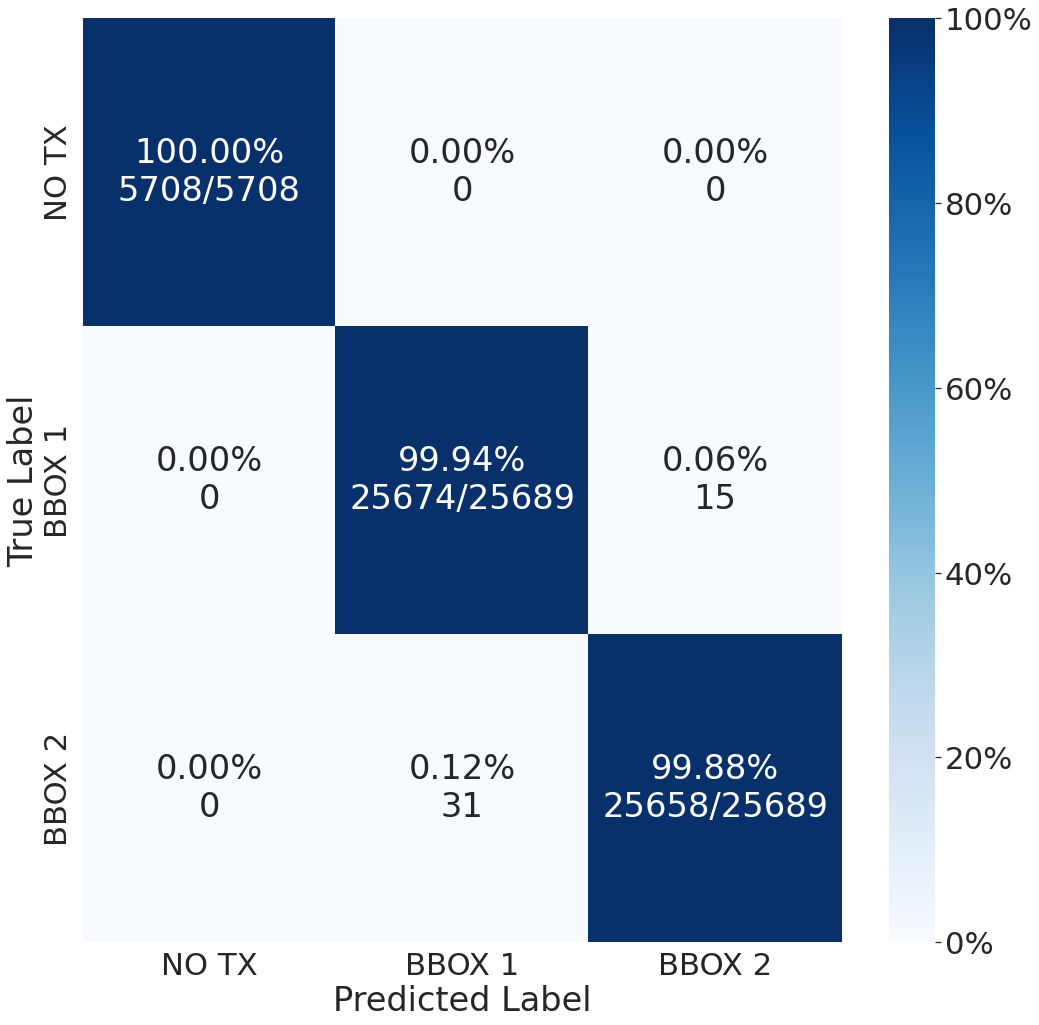}
	\caption{Confusion Matrix for Setup 1 data trained with Neural Network Classifier.}
	\label{fig:cf_setup1_nn}
\end{figure}
%%%%%%%%%%%%%
% END FIGURE
%%%%%%%%%%%%%

%%%%%%%%%%%%%
% FIGURE
%%%%%%%%%%%%%
\begin{figure}[h!]
	\centering
	\includegraphics[scale=\sizeConfusionMatrix]{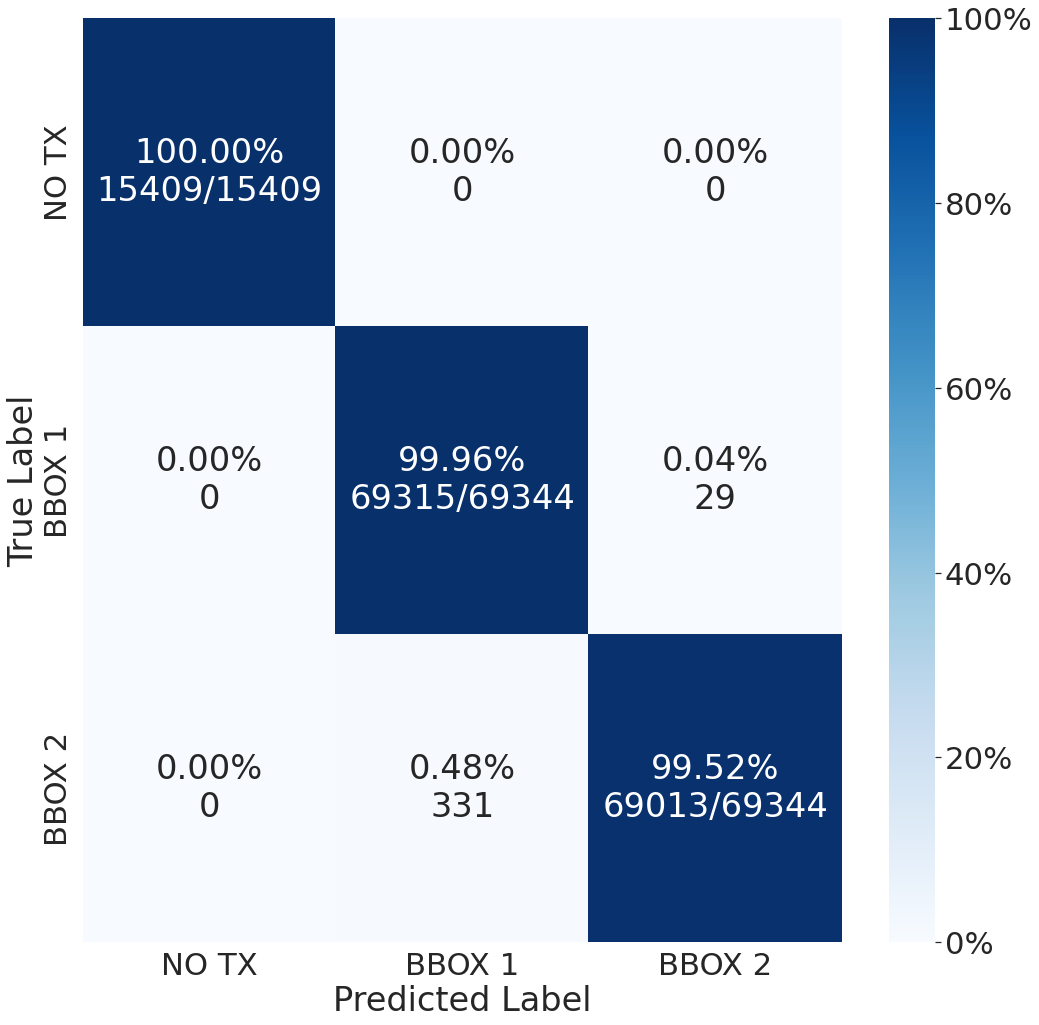}
	\caption{Confusion Matrix for Setup 2 data trained with Random Forest Classifier.}
	\label{fig:cf_setup2_rfc}
\end{figure}
%%%%%%%%%%%%%
% END FIGURE
%%%%%%%%%%%%%

%
%\Figure[hbtp](topskip=-1pt, botskip=0pt, midskip=0pt)[width=0.4\textwidth]{imgs/results/cm/setup2_nn/cf_setup2_nn}
%{Confusion Matrix for Setup 2 data trained with Neural Network Classifier.\label{fig:cf_setup2_nn}}

%%%%%%%%%%%%%
% FIGURE
%%%%%%%%%%%%%
\begin{figure}[h!]
	\centering
	\includegraphics[scale=\sizeConfusionMatrix]{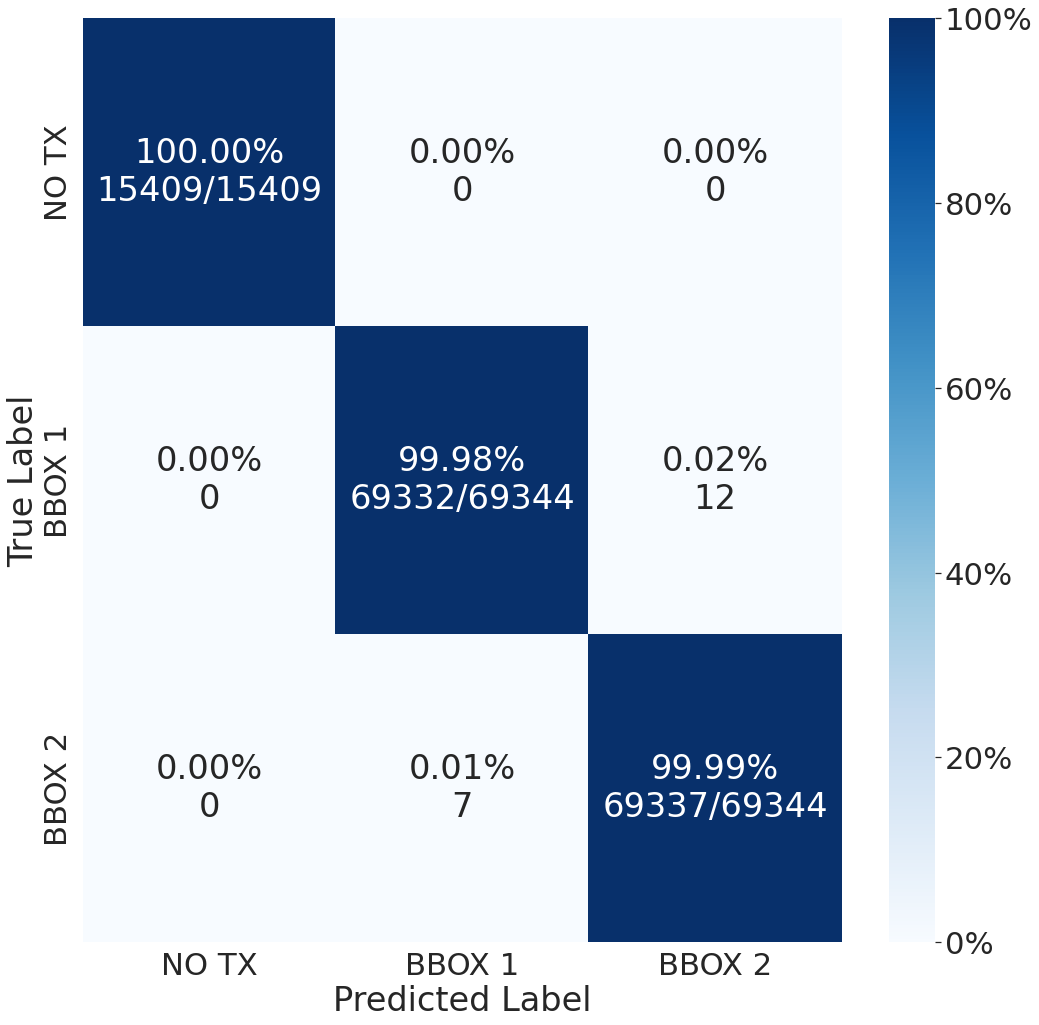}
	\caption{Confusion Matrix for Setup 2 data trained with Neural Network Classifier.}
	\label{fig:cf_setup2_nn}
\end{figure}
%%%%%%%%%%%%%
% END FIGURE
%%%%%%%%%%%%%

%%%%%%%%%%%%%
% FIGURE
%%%%%%%%%%%%%
\begin{figure}[h!]
	\centering
	\includegraphics[scale=\sizeConfusionMatrix]{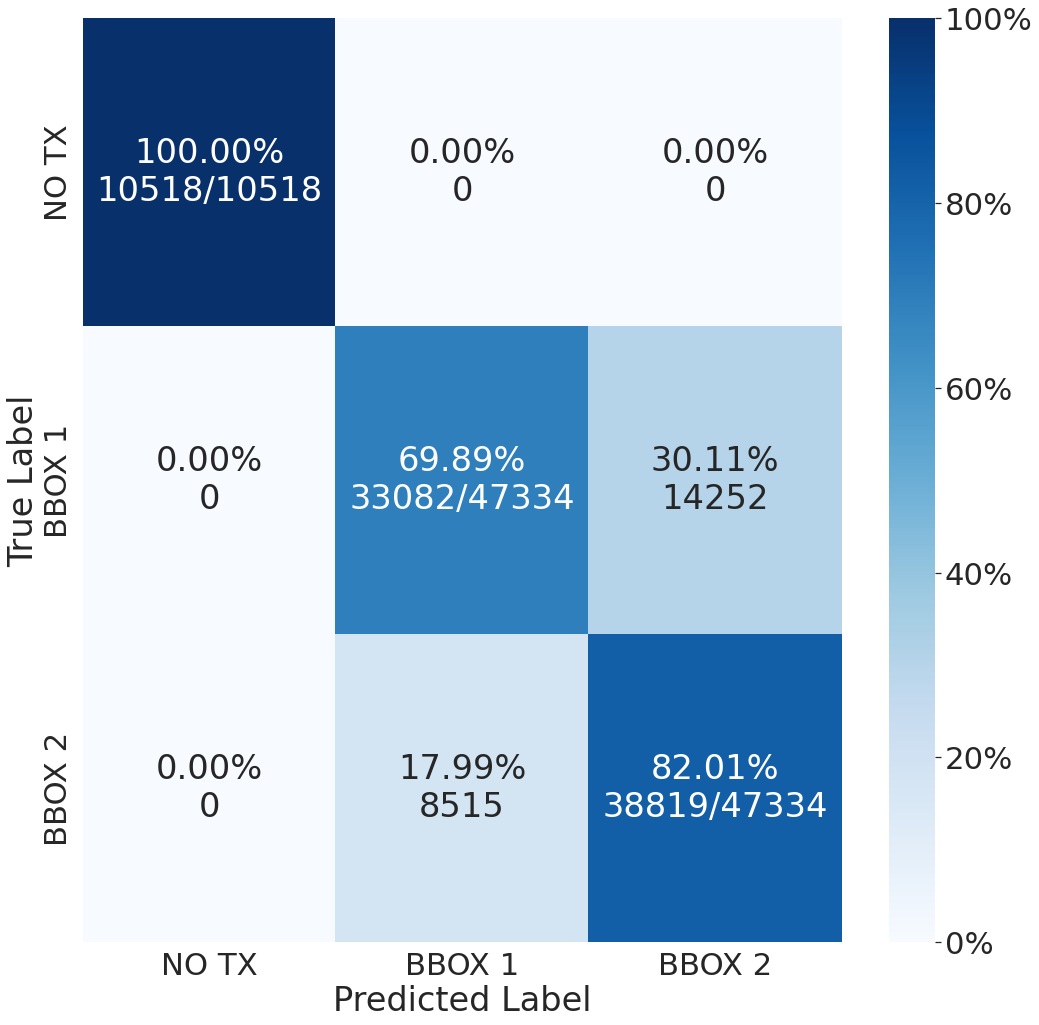}
	\caption{Confusion Matrix for Setup 3 data trained with Random Forest Classifier.}
	\label{fig:cf_setup3_rfc}
\end{figure}
%%%%%%%%%%%%%
% END FIGURE
%%%%%%%%%%%%%

%%%%%%%%%%%%%
% FIGURE
%%%%%%%%%%%%%
\begin{figure}[h!]
	\centering
	\includegraphics[scale=\sizeConfusionMatrix]{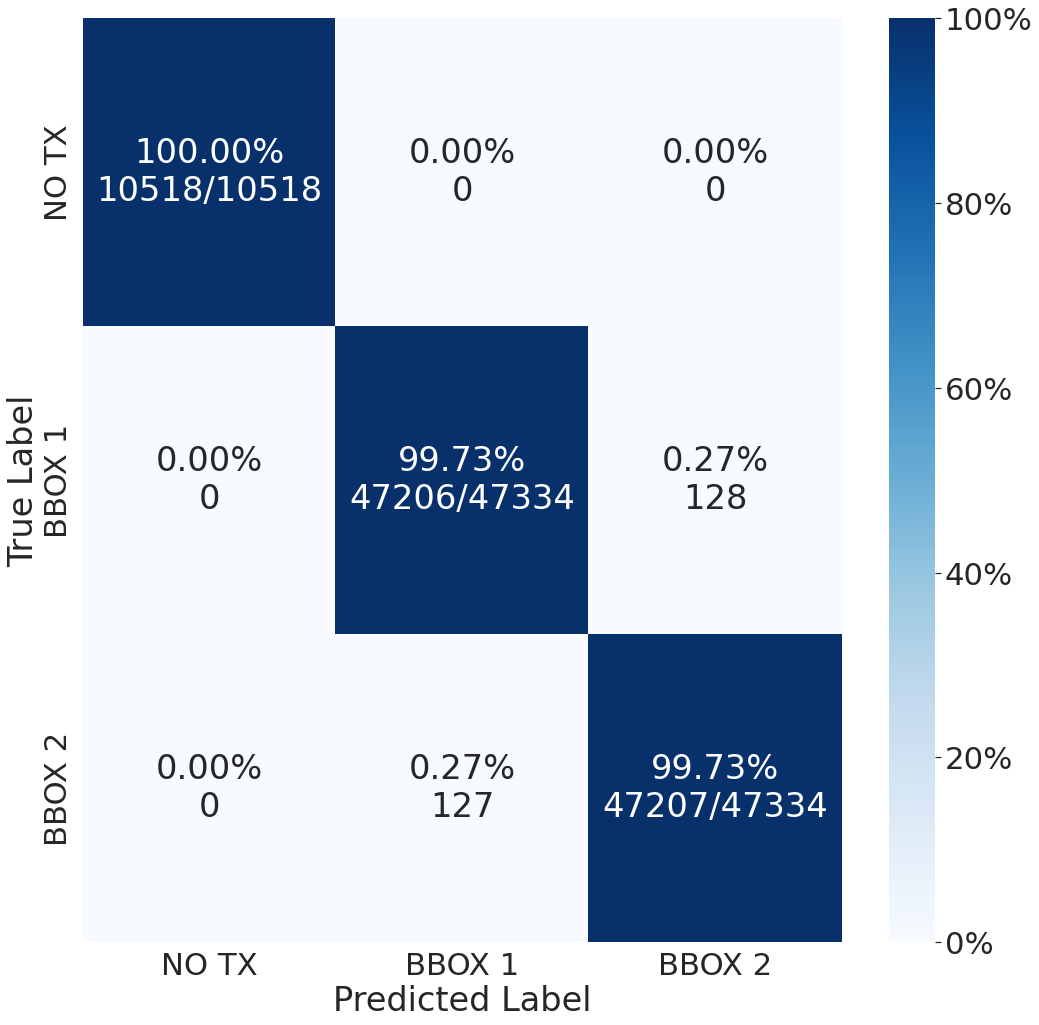}
	\caption{Confusion Matrix for Setup 3 data trained with Neural Network Classifier.}
	\label{fig:cf_setup3_nn}
\end{figure}
%%%%%%%%%%%%%
% END FIGURE
%%%%%%%%%%%%%

%%%%%%%%%%%%%
% FIGURE
%%%%%%%%%%%%%
\begin{figure}[h!]
	\centering
	\includegraphics[scale=\sizeConfusionMatrix]{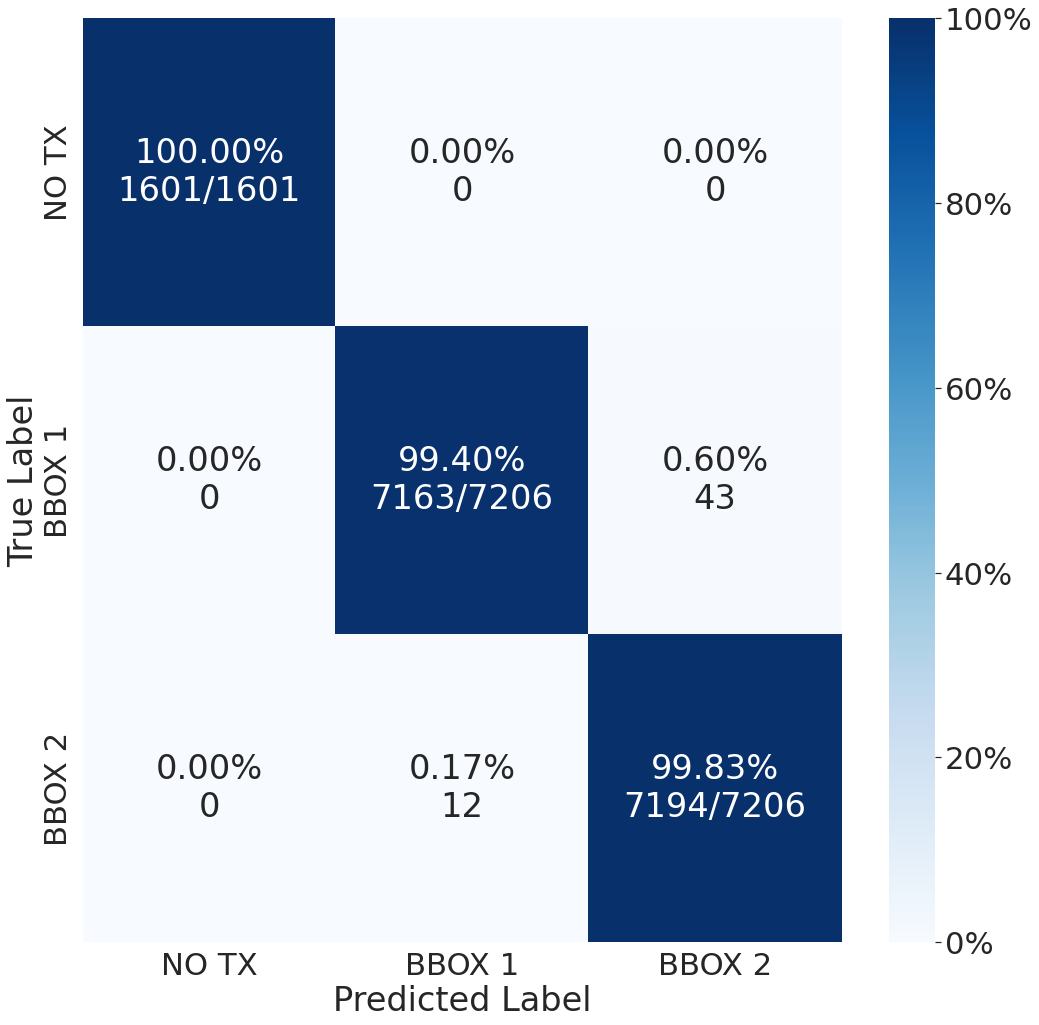}
	\caption{Confusion Matrix for Setup 4 data trained with Random Forest Classifier.}
	\label{fig:cf_setup4_rfc}
\end{figure}
%%%%%%%%%%%%%
% END FIGURE
%%%%%%%%%%%%%

%%%%%%%%%%%%%
% FIGURE
%%%%%%%%%%%%%
\begin{figure}[h!]
	\centering
	\includegraphics[scale=\sizeConfusionMatrix]{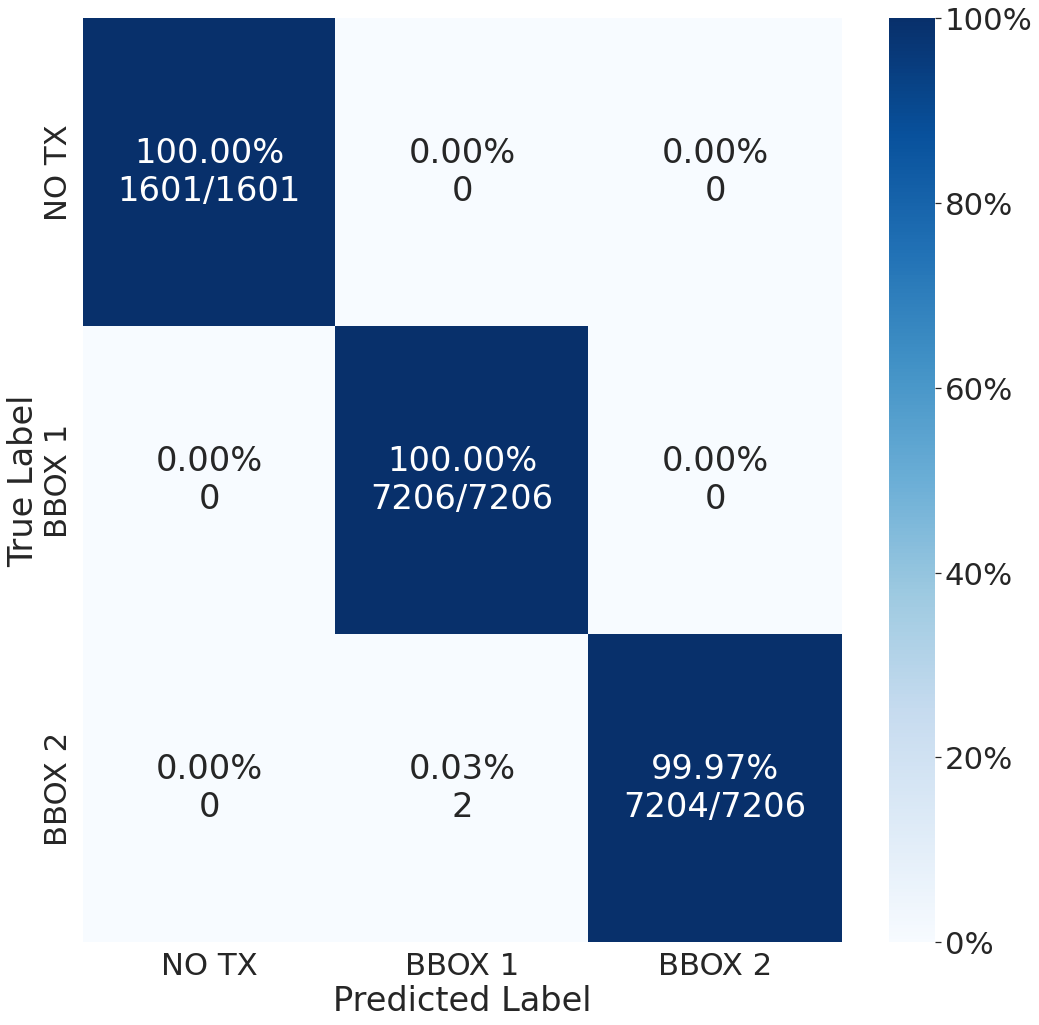}
	\caption{Confusion Matrix for Setup 4 data trained with Neural Network Classifier.}
	\label{fig:cf_setup4_nn}
\end{figure}
%%%%%%%%%%%%%
% END FIGURE
%%%%%%%%%%%%%

%%                          

The training time for the experiments using neural network classifiers was on average $3$ times lower than the ones with random forest classifiers. The longer training duration occurs because the random forest included an exhaustive grid search for parameters and cross-validation during training. The neural network classifiers were training during $10$ epochs and no hyper-parameter search was used. Hence the smaller training time for the classifiers using neural networks.

The performance metrics show that experiments with the random forest classifiers had $F_{1}$-scores equal to or higher than $0.84$. This is also true for precision and recall. These results still give us a precise and robust classifier; it correctly classifies the instances, even if they are difficult to classify. These numbers are from the Setup 3, with the largest training dataset. The reason for lower performance metrics, when compared to the other experiments, can be found in the search space used for hyperparameter tuning. The numbers of trees and tree depth presented in Section~\ref{subsec:rand-forest} did not contain the hyper-parameter values needed for this experiment to succeed. A better solution can be found with a greater number of trees in the ensemble. With $227$ trees and tree depth of $65$ the $F_{1}$-scores is $0.97$. However, more trees in the ensemble increases the time necessary for the model to make a classification in the deployment phase. In this work, we maintained the same search space in all the experiments to make the comparisons fair. 

In a practical case, the search space for the \gls{rfc} can be changed until the best solution is found. The training duration is in the order of minutes, hence it is feasible to train multiple times for the same set of measurements. After the training phase, during the deployment phase the model gives an output in a negligible amount of time. In this sense, the cost of retraining the dataset is not high, even for the random forest classifiers.

The experiments with the neural network classifiers achieved $F_{1}$-score of $0.99$ in every setup. Only a minor part of the dataset instances were incorrectly classified. With a small architecture of the neural networks, as displayed in Table~\ref{tab:nn-arch}, the models can train fast and still excel in the classification task, as shown by the performance metrics. 

Overall, the high accuracy and $F_{1}$-score in the experiments show the capability of the proposed framework to perform well across different environments. Using the testbed described in Section~\ref{subsec:scenario}, we tested the proposed framework using datasets with different sizes, collected in different types of places. The results confirm that our solution is capable of correctly match the user identity in a video feed with its corresponding radio signal. 

The experimental testbed described in Section~\ref{subsec:scenario} can then be further extended. It is possible to use our proposed framework to include more devices in the scene. Although not strictly necessary, it is possible to use more cameras to capture different angles of the environment. The framework is flexible to adapt and work in more realistic scenarios. For example, we can use two different cameras to detect four possible users at the same time. The input instance for the classifier from Figure~\ref{fig:input_class} would have the CIR, the CIR-related features for the radio data. For the video feed features with two cameras, C1 and C2, we would have ``BBOX 1 - C1'' through ``BBOX 4 - C2''. Each camera contributes with a bounding box indicating the position of the user in the scene. There would be five different classes that the classifier would be trained on. As the features are an array of numbers, with more cameras and possible transmitting-users it does not increase the number of training data as using the whole image for training. This makes our solution scalable to more complex scenarios.

%%%%%%%%%%%% CONCLUSIONS %%%%%%%%%%%%
%%%%%%%%%%%%%%%%%%%%%%%%%%%%%%%%%%%%%%%%%%%%%%%%%%%
% SECTION: CONCLUSIONS
%%%%%%%%%%%%%%%%%%%%%%%%%%%%%%%%%%%%%%%%%%%%%%%%%%%
\section{Conclusions}
\label{sec:conclusions}
This work described the procedures for the integration of a computer vision system with a radio access network through means of artificial intelligence. Our work showcases the identification of the true radio transmitter between two equipment existing in a video feed. We showed that by modeling the problem as a classification task and using machine learning techniques, random forest and deep neural network classifiers, we were able to correctly identify the true transmitter in the scene in several different scenarios presented. 
We carried out experiments using measurements collected in four different environments. The performance metrics computed show the proposed solution is capable of correctly identifying the users with very high accuracy in all tested environments. The proposed framework was shown to be very robust and reliable yet flexible. It is possible to extend the testbed used here for a proof-of-concept and experiment with more realistic scenarios.

This work is a building block for the integration of different sensors for the improvement of context-aware communication systems. This integration is going to be ubiquitous in the following generations. For this reason, our solution can be used in other projects working with joint technologies. Industrial private networks can take advantage of this integration. Since the users are mainly robots belonging to the company there are no privacy issues, allowing the extraction of useful data from visual sources. 

%%%%%%%%%%%% BIBLIOGRAPHY %%%%%%%%%%%%

\bibliographystyle{template/splncs04}
\bibliography{main}

\end{document}